\documentclass[sn-mathphys,Numbered]{sn-jnl}

\usepackage{subcaption}
\usepackage{multirow}%
\usepackage{pgffor}%
\usepackage{amsmath,amssymb,amsfonts}%
\usepackage{amsthm}%
\usepackage{mathrsfs}%
\usepackage[inkscapelatex=false]{svg}
\usepackage[title]{appendix}%
\usepackage{xcolor}%
\usepackage{textcomp}%
\usepackage{booktabs}%
\usepackage{manyfoot}%
\usepackage{booktabs}%
\usepackage{tabularx}%
\usepackage{algorithm}%
\usepackage{adjustbox}
\usepackage{comment}
\usepackage{algorithmicx}%
\usepackage{algpseudocode}%
\usepackage{listings}%
\usepackage{caption}%
\usepackage{array}%
\usepackage{lineno}%

\usepackage{float}%

\theoremstyle{thmstyletwo}%
\theoremstyle{thmstylethree}%
\usepackage{titlesec}

\begin{document}

\title[Article Title]{A Generalist Model for Diverse Text-Guided Medical Image Synthesis}

\author[1]{\fnm{Cho} \sur{Joseph}}\email{jcho5@stanford.edu}
\equalcont{These authors contributed equally to this work.}
\author[1]{\fnm{Mathur} \sur{Mrudang}}
\equalcont{These authors contributed equally to this work.}
\author[1]{\fnm{Zakka} \sur{Cyril}}
\equalcont{These authors contributed equally to this work.}
\author[1]{\fnm{Kaur} \sur{Dhamanpreet}}
\author[1]
{\fnm{Leipzig} \sur{Matthew}}
\author[1]
{\fnm{Dalal} \sur{Alex}}
\author[1]
{\fnm{Krishnan} \sur{Aravind}}
\author[2]{\fnm{Koo} \sur{Eubee}}
\author[2]{\fnm{Wai} \sur{Karen}}
\author[2]{\fnm{Zhao S.} \sur{Cindy}}
\author[5, 6]{\fnm{Chaudhari} \sur{Akshay}}
\author[1]
{\fnm{Duda} \sur{Matthew}}
\author[1]
{\fnm{Choi} \sur{Ashley}}
\author[2]
{\fnm{Rahimy} \sur{Ehsan}}
\author[2]
{\fnm{Azzouz} \sur{Lyna}}
\author[1]{\fnm{Fong} \sur{Robyn}}
\author[3]{\fnm{Shad} \sur{Rohan}}
\author[1]{\fnm{Hiesinger} \sur{William}}\email{willhies@stanford.edu}

\affil[1]{\orgdiv{Department of Cardiothoracic Surgery}, \orgname{Stanford Medicine}}

\affil[2]{\orgdiv{Department of Ophthalmology}, \orgname{Stanford Medicine}}

\affil[3]{\orgdiv{Division of Cardiovascular Surgery}, \orgname{Penn Medicine}}

\affil[5]{\orgdiv{Department of Radiology}, \orgname{Stanford Medicine}}
\affil[6]{\orgdiv{Department of Biomedical Data Science}, \orgname{Stanford Medicine}}

\abstract{
Deep learning algorithms require extensive data to achieve robust performance. However, data availability is often restricted in the medical domain due to patient privacy concerns. Synthetic data presents a possible solution to these challenges. Image generative models have found increasing use for medical applications, but are often task-specific, thus limiting their scalability. Moreover, existing models frequently rely on private datasets for training, which constrain their reproducibility. To address this, we introduce MediSyn: an open-access, generalist, text-guided latent diffusion model capable of generating synthetic images across 6 medical specialties and 10 imaging modalities, while being trained exclusively on publicly available data. Through extensive experimentation, we provide several key contributions. First, we demonstrate that training a generative model on visually diverse medical images does not degrade synthetic image quality. Second, we show that this generalist approach is substantially more computationally efficient than a coordinated suite of task-specific models. Third, we establish that a generalist model can produce realistic, text-aligned synthetic images across visually and medically distinct modalities, as validated by expert physicians. Fourth, we provide empirical evidence that these synthetic images are visually distinct from their corresponding real patient images, alleviating concerns about data memorization in image generative models. Finally, we demonstrate that a generalist model can produce synthetic images that improve classifier performance in data-limited settings across multiple medical specialties. Altogether, our findings highlight the immense potential of generalist image generative models to accelerate algorithmic research and development in medicine.}

\maketitle

\renewcommand{\figurename}{Fig}
\begin{figure}[H] 
    \centering
    \includegraphics[width=\linewidth]{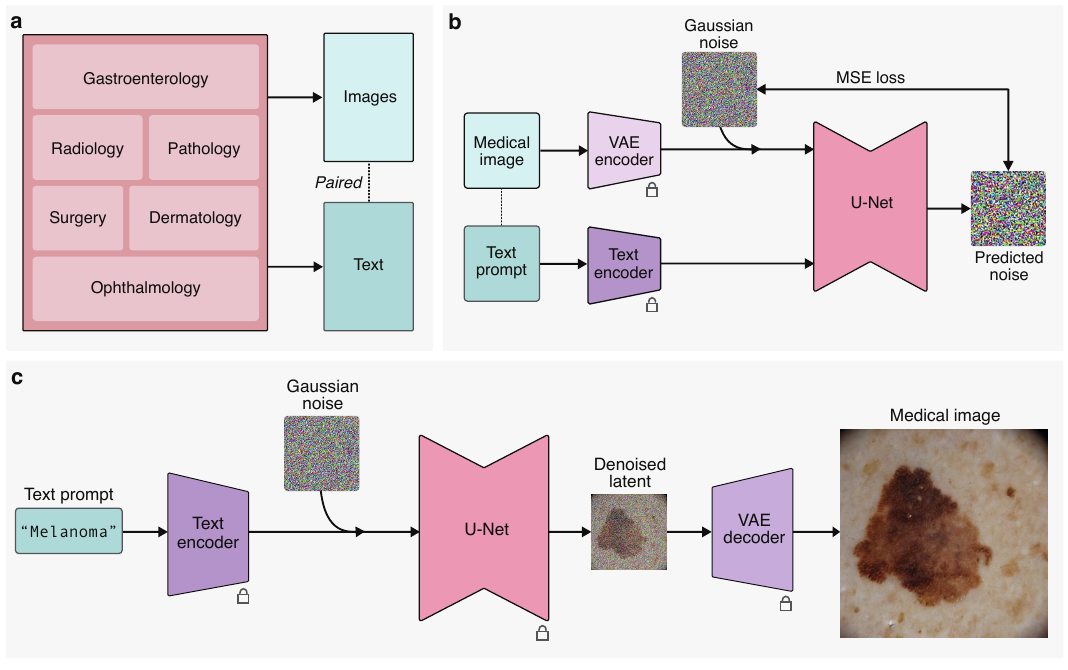} 
    \caption{\textbf{Overview of MediSyn framework.} \textbf{a.} Training dataset: A large-scale corpus of medical image-text pairs across 6 medical specialties was curated from the public domain. \textbf{b.} Training procedure: A text-conditional U-Net is trained to denoise a latent space representation of an image intentionally corrupted with Gaussian noise. \textbf{c.} Inference procedure (text-to-image generation): The trained U-Net progressively denoises a latent vector sampled from a Gaussian distribution, which is then decoded by a variational autoencoder (VAE) into a high-quality synthetic medical image. MSE: mean squared error.
 }
    \label{fig:overview_figure}
\end{figure}

\section{Introduction}\label{sec1}
Deep learning (DL) has made remarkable strides in medicine, with applications ranging from diagnostic imaging to predictive analytics \cite{pancreatic, timediabetic, biomarker, kline2022multimodal, Li2023}. Despite these advances, concerns about patient privacy preclude the large-scale adoption and development of DL models in medicine \cite{lavanchy2023preserving, nwoye2024surgical}. While large volumes of clinical data are generated by the healthcare industry worldwide, these are often withheld from public release to comply with patient privacy guidelines. As such, these data remain unavailable for algorithmic research and development in medicine \cite{Yadav2023}. To overcome this, various groups have publicly released de-identified medical datasets, in which protected health information is scrambled or removed to prevent patient identification \cite{Johnson2019}. However, current de-identification techniques are often tedious, cost-prohibitive, and susceptible to re-identification attacks \cite{Packhauser2022}. While various techniques such as federated learning \cite{Rauniyar2022FederatedLF} and differential privacy \cite{Dwork2011} have been proposed to mitigate privacy risks, they often compromise model performance and may also be susceptible to re-identification attacks \cite{Enthoven2020Overview, Bhanbhro2024, bagdasaryan2019differential}. In light of these challenges, most medical data remain privately held and used. Since DL systems are data-intensive, the issue of data scarcity can inhibit the performance of DL-based clinical support systems \cite{DDI, multimodal}.

\renewcommand{\figurename}{Fig}
\begin{figure}[] 
    \centering
     \includegraphics[width=\linewidth]{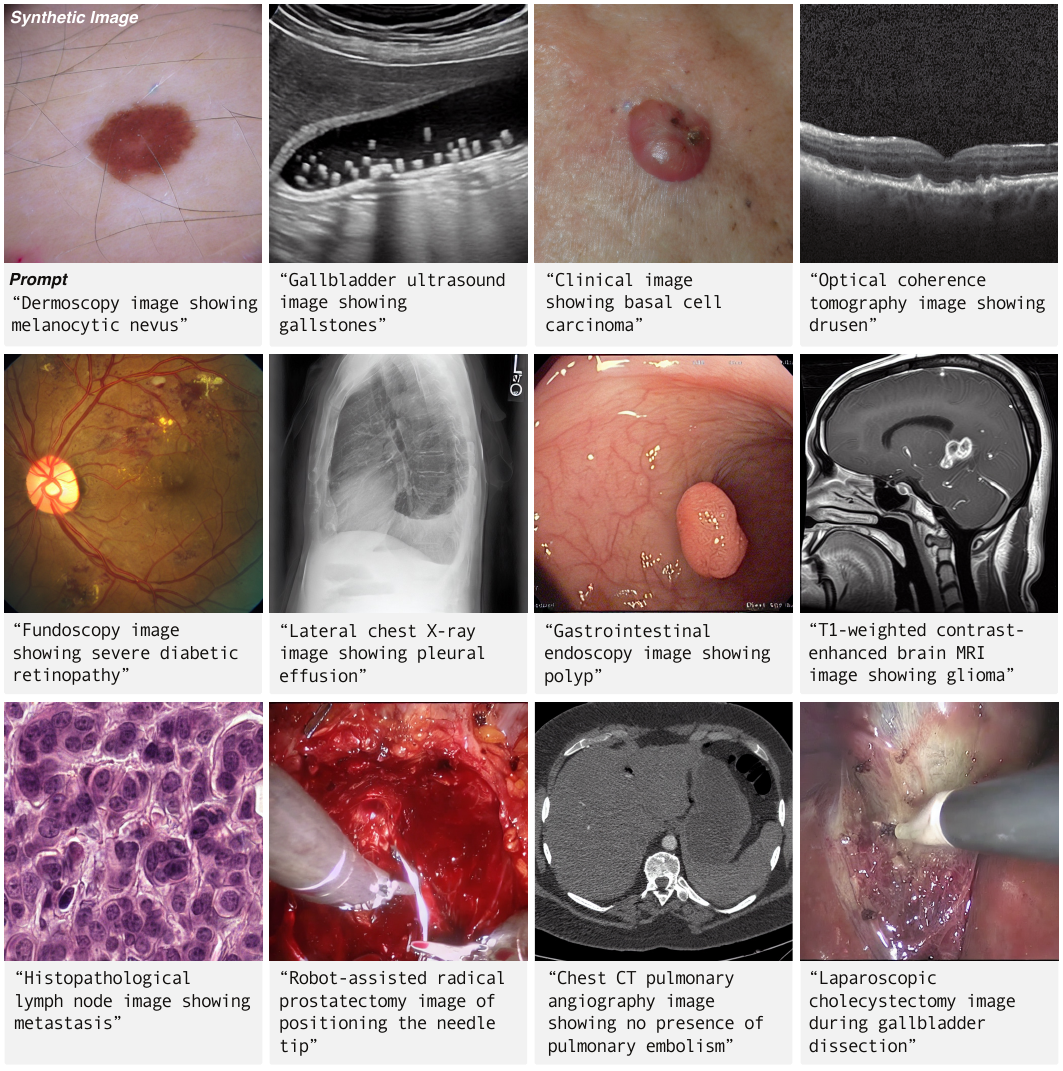} 
    \caption{\textbf{Text-conditioned synthesis of medical images.} A series of synthetic images generated by MediSyn, across 6 medical specialties and 10 imaging modalities. The accompanying captions served as the text prompts for our model.
 }
    \label{fig:generated_images}
\end{figure}

In recent years, image generative models have emerged as a promising solution that preserves patient privacy \cite{Schutte2021, pathLDM}. To do so, such models are trained to learn the overall statistical distributions of training data instead of memorizing specific images. Thus, they may be used to generate high-quality data for algorithmic research and development in medicine \cite{vanBreugel2024synthetic}. In particular, latent diffusion models (LDMs) have received great interest due to their ability to efficiently generate realistic and diverse synthetic images, achieving state-of-the-art results in output quality metrics \cite{Rombach_2022_CVPR, SDXL}. In addition, they have been adapted to incorporate text prompts that enable precise control over the image generation process \cite{zhang2024text}. 

Within medicine, LDMs have been adapted to generate images in specialties such as radiology \cite{ROENT, LiverDiff}, pathology \cite{pathLDM, pathdiff_scireports}, and dermatology \cite{SagersDERM, kim2025diffusion}. Despite these advances, existing models remain largely task-specific, restricted to a single medical specialty, imaging modality, and anatomical region. Consequently, separate image generative models must be trained, deployed, and maintained for each medical data type, substantially increasing computational and operational burden \cite{diffmedsurvey, implementation}. This burden is further amplified by the significant GPU resources required for diffusion models, which remain inaccessible to many hospital systems seeking to support a range of such systems \cite{LDM, Jia2023ResourceAwarenessAIHealthcare}. In addition, existing medical image generative models frequently rely on private datasets for training, which limit their reproducibility \cite{SHIFTS, MINIM, MedSyn, dai2025improving, BUSGen}. To overcome this challenge, we present MediSyn – a singular, text-guided LDM for image synthesis across 6 medical specialties and 10 imaging modalities. To train MediSyn in a unified manner, we curated a diverse, publicly sourced corpus of 1,260,826 medical images paired with detailed text descriptions. Next, across five distinct assessments of MediSyn, we establish several key findings. First, we demonstrate that training a generative model on visually diverse medical images does not degrade synthetic image quality. Second, we show that this generalist approach is substantially more computationally efficient than a coordinated suite of task-specific models. Third, we establish that a generalist model can produce realistic, text-aligned synthetic images across visually and medically distinct modalities, as validated by expert physicians. Fourth, we provide empirical evidence that these synthetic images are visually distinct from their corresponding real patient images, alleviating widespread concerns about data memorization in image generative models. Finally, we demonstrate that a generalist model can produce synthetic images that improve classifier performance in data-limited settings across multiple medical specialties.

\section{Results}\label{results}

To create a generalist, text-guided latent diffusion model for diverse medical image synthesis, we fine-tuned Stable Diffusion version 1.4 \cite{Rombach_2022_CVPR} on over 1 million publicly available medical images, see Figure \ref{fig:overview_figure}. Our trained model --  ``MediSyn," can generate synthetic medical images across 6 medical specialties (Gastroenterology, Radiology, Pathology, Surgery, Dermatology, and Ophthalmology) and 10 imaging modalities (Computed Tomography,
X-ray, Magnetic Resonance Imaging, Ultrasound, Endoscopy, Microscopy, Fundoscopy,
Optical Coherence Tomography, Dermoscopy, and Clinical images), see Figure \ref{fig:generated_images}. 

In this study, we provide several key contributions through five distinct evaluations of MediSyn. First, we show that training a generalist model on visually diverse medical images does not degrade synthetic image quality. To do so, we quantitatively compare the fidelity and diversity of images generated by MediSyn against those generated by task-specific models. Second, we demonstrate that this generalist approach is more computationally efficient than a coordinated suite of task-specific models. To do so, we compared the inference-time costs of generating diverse medical images with MediSyn against a coordinated suite of task-specific models. Third, we establish that a generalist model can produce realistic, text-aligned synthetic images across visually and medically distinct modalities, as validated by expert physicians. To do so, we conduct a ``visual Turing test" and a blinded text-alignment evaluation with physicians trained in general surgery, cardiothoracic surgery, and ophthalmology. Fourth, we provide empirical evidence that these synthetic images are visually distinct from their corresponding real patient images, alleviating concerns about data memorization in image generative models. To do so, we performed pairwise comparisons between MediSyn-generated images and their nearest neighbors in the training data of real patient images. Finally, we demonstrate that a generalist model can produce synthetic images that improve classifier performance in data-limited settings across multiple medical specialties. To do so, we train classifiers across multiple medical specialties on real data, MediSyn-generated data, or real data supplemented with MediSyn-generated data, and compare performance at various data ratios.

\subsection{Diverse medical image training does not degrade synthetic image quality}
We hypothesize that training a generative model on visually diverse medical images does not degrade synthetic image quality. To test this, we compared the Fréchet Inception Distance (FID) \cite{heusel2017gans} and Multi-scale Structural Similarity Index Metric (MS-SSIM) \cite{MS-SSIM} scores between MediSyn and task-specific models, similar to prior works \cite{ROENT, MINIM}. We perform this evaluation across three visually and medically distinct imaging modalities: musculoskeletal X-ray, dermoscopy images, and images from robot-assisted radical prostatectomy. The task-specific models were trained on datasets restricted to a single medical specialty, imaging modality, and anatomical region, see Table \ref{tab:GenvSpec}. FID was computed as the distance between each generative model's outputs and its corresponding real training data distribution, before FID scores were compared across models. MS-SSIM was calculated only among generated images from a given model before MS-SSIM values were similarly compared across models. Lower FID scores indicate that synthetic images more closely mirror the corresponding real training data distribution (higher fidelity), whereas lower MS-SSIM scores indicate greater image diversity.

For both dermoscopy and robot-assisted radical prostatectomy images, MediSyn achieved lower FID but higher MS-SSIM scores than its task-specific counterparts. Thus, images from MediSyn may better match the distribution of real data but are less diverse. In contrast, MediSyn produced a higher FID but a lower MS-SSIM than the task-specific model for musculoskeletal x-ray images. Together, these results suggest that a generalist model yields comparable image fidelity and diversity when compared to task-specific models.

 \begin{table}[t!]
    \centering
    \caption{Quantitative evaluation of MediSyn and task-specific image generative models trained on datasets restricted to a single medical specialty, imaging modality, and anatomical region. Task-specific models are denoted by SD-\{image-type\}.  Lower Fréchet inception distance (FID) indicates the generated image distribution more closely aligns with the original image distribution. Lower multi-scale structural similarity index metric (MS-SSIM) values indicate greater diversity among the synthetic images. SD: Stable Diffusion; MSK: musculoskeletal; RARP: robot-assisted radical prostatectomy.}
    \begin{tabular}{>{\raggedright\arraybackslash}m{0.3\linewidth} 
                    >{\raggedright\arraybackslash}m{0.2\linewidth} 
                    >{\raggedleft\arraybackslash}m{0.2\linewidth} 
                    >{\raggedleft\arraybackslash}m{0.2\linewidth}}
        \toprule
        \textbf{Image type} & \textbf{Model} & \textbf{FID} ($\downarrow$) & \textbf{MS-SSIM} ($\downarrow$) \\
        \midrule
        {MSK X-ray} & MediSyn  & 78.87 & $\mathbf{0.21 \pm 0.16}$ \\
         & SD-MSK X-ray  & \textbf{74.49} & $0.28 \pm 0.15$ \\
        \midrule
        {Dermoscopy}  & MediSyn  & \textbf{56.50} & $0.49 \pm 0.14$\\
         & SD-Dermoscopy  & 83.59 & $\mathbf{0.35 \pm 0.20}$ \\
        \midrule
        {RARP}  & MediSyn  & \textbf{18.28} & $0.13 \pm 0.08$ \\
         & SD-RARP  & 21.14 & $\mathbf{0.10 \pm 0.06}$ \\
        \bottomrule
    \end{tabular}
    
    \label{tab:GenvSpec}
\end{table}

\subsection{Generalist models are more computationally efficient than a task-specific suite}
We hypothesize that our generalist approach provides a more computationally efficient approach to generating diverse medical images than a coordinated suite of task-specific models. To test this, we measured and compared inference-time costs for generating 10,000 images across 6 medical specialties and 10 imaging modalities between MediSyn and the task-specific suite, see Table \ref{tab:benchmark}. Please note that MediSyn and the task-specific models share the same underlying architecture, with the task-specific models restricted to a single medical specialty, imaging modality, and anatomical region.

To develop the task-specific suite, we implemented a rule-based router that assigns each input text prompt to a corresponding task-specific model based on its content. At inference time, if a text prompt required a different model than the preceding prompt, the corresponding task-specific model weights were dynamically loaded into a shared U-Net instance, after which image generation proceeded in the standard manner. In contrast, MediSyn used a single shared U-Net with fixed weights capable of handling text prompts spanning multiple medical disciplines, eliminating the need for prompt routing or model switching at inference time. 

When averaged over 5 runs, MediSyn generated 10,000 images in $4.34 \pm 0.04$ hours while the task-specific suite did so in $9.18 \pm 1.12$ hours. In detail, MediSyn required a longer initialization time for the full Stable Diffusion–based pipeline ($9537.20 \pm 4607.61$ ms) when compared to the task-specific suite ($3678.35 \pm 638.07$ ms). This stems from an additional step in which the default Stable Diffusion U-Net weights are replaced with MediSyn's fine-tuned weights. Due to their shared architecture, image generation time for MediSyn and the task-specific suite was nearly identical; $1562.75 \pm 14.34$ ms and $1559.36 \pm 12.61$ ms per image, respectively.  However, the task-specific suite incurred substantial overhead from loading model weights from CPU memory into the shared GPU-resident model. This resulted in an additional cost of $1914.88 \pm 468.16$ ms per model switch, which exceeded the cost of image generation itself. In contrast, routing overhead was negligible at $0.0024 \pm 0.0001$ ms per route, and loading task-specific model weights into CPU memory incurred a modest cost of $73.54 \pm 9.61$ ms per model switch. Notably, MediSyn outperforms the task-specific suite when generating as few as approximately 3 images, see Supplementary Figure S2. Together, these results demonstrate that a generalist model achieves substantially greater computational efficiency than a task-specific suite and represents a more scalable approach for synthetic medical image generation.

\begin{table}[t!]
    \caption{Comparison of MediSyn and a task-specific model suite in computational efficiency for diverse medical image generation. Model initialization denotes the time required to load the full Stable Diffusion-based model pipeline (text encoder, variational autoencoder, and U-Net). Routing (ms) denotes the time required to assign each text prompt to a corresponding task-specific model. CPU loading (ms) and GPU loading (ms) correspond to the time required to load task-specific U-Net weights into CPU memory and to transfer them into a shared GPU-resident U-Net, respectively. Image generation (ms) measures the time required to generate an image once model weights are loaded. Total time: 10k imgs (h) represents the total time required to generate all 10,000 images. Values are reported as mean $\pm$ standard deviation across five runs.}
    \begin{tabular}{>{\raggedright\arraybackslash}m{0.35\linewidth}
                    >{\raggedleft\arraybackslash}m{0.25\linewidth}
                    >{\raggedleft\arraybackslash}m{0.25\linewidth}}
        \toprule
        \textbf{Stage-wise Timing} & \textbf{MediSyn} & \textbf{Task-specific Suite} \\
        \midrule
        Model initialization (ms) & 9537.20 $\pm$ 4607.61 & 3678.35 $\pm$ 638.07 \\
        Routing (ms) & -- & 0.0024 $\pm$ 0.0001 \\
        CPU loading (ms) & -- & 73.54 $\pm$ 9.61 \\
        GPU loading (ms) & -- & 1914.88 $\pm$ 468.16 \\
        Image generation (ms) & 1562.75 $\pm$ 14.34 & 1559.36 $\pm$ 12.61 \\
        \midrule
        \textbf{Total time: 10k images (h) } & \textbf{4.34 $\pm$ 0.04} & \textbf{9.18 $\pm$ 1.12} \\  
        \bottomrule
    \end{tabular}
    \label{tab:benchmark}
\end{table}

\subsection{Generalist models can produce realistic, text-aligned images across medically distinct modalities}
\begin{figure}[H] 
    \centering
    \includegraphics[width=\linewidth]{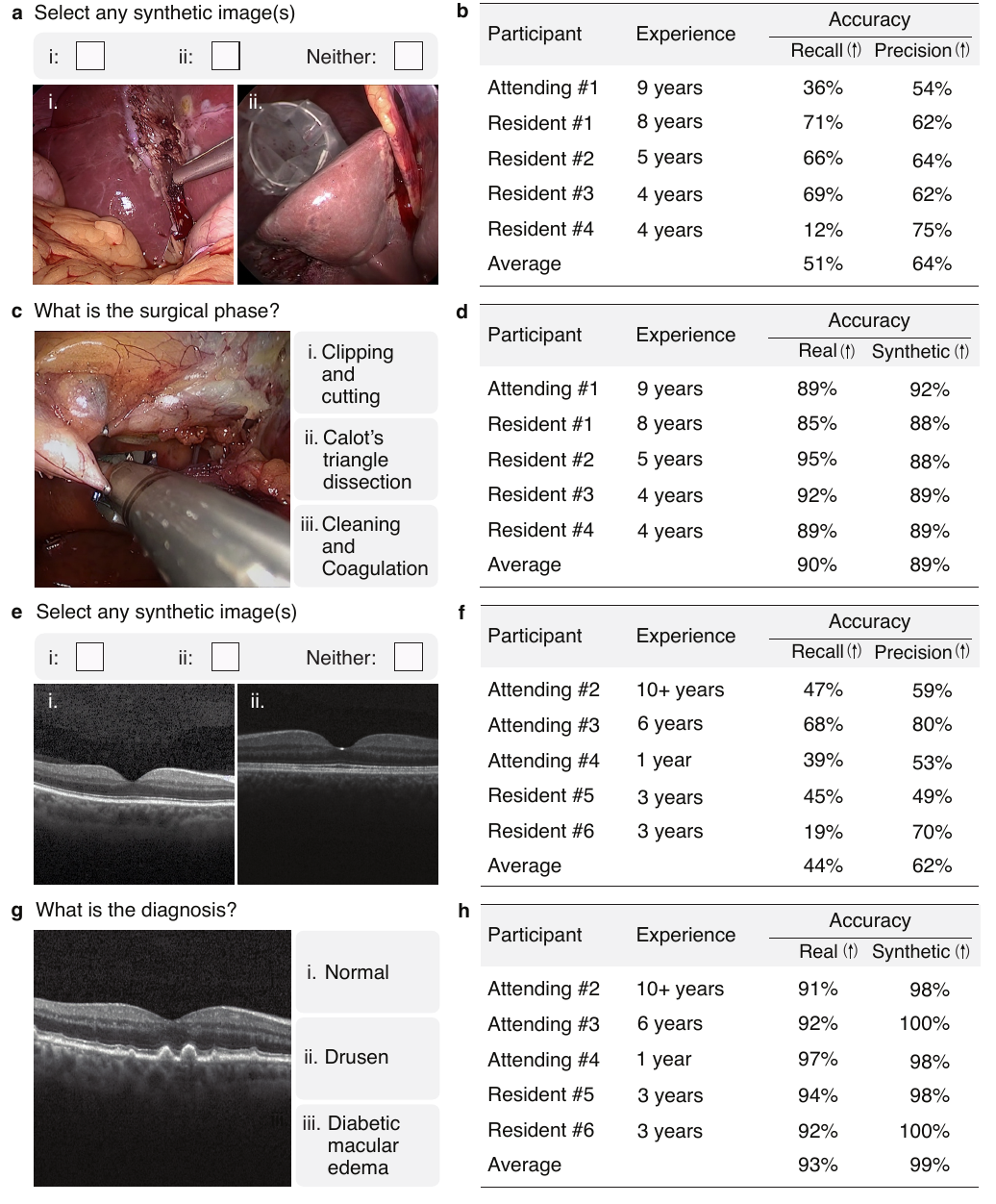} 
    \caption{\textbf{Physician assessments of synthetic and real medical images.} \textbf{a.} Surgeons were asked to select any synthetic image(s) in image pairs of laparoscopic cholecystectomy. \textbf{b.} Surgeon performance while identifying synthetic surgical images. \textbf{c.} Additionally, surgeons were asked to classify the surgical phase for a set of real and synthetic laparoscopic cholecystectomy images. \textbf{d.} Surgeon performance on the surgical image classification task. 
    \textbf{e.} Ophthalmologists were asked to select any synthetic image(s) in image pairs of optical coherence tomography images. \textbf{f.} Ophthalmologist performance while identifying synthetic optical coherence tomography images. \textbf{g.} Additionally, ophthalmologists were asked to classify the disease condition for a set of real and synthetic optical coherence tomography images. \textbf{h.} Ophthalmologist performance on the optical coherence tomography image classification task. Please note that experience is measured as time in current position. Metrics are annotated with an upward arrow to indicate that higher values reflect better physician performance. }
    \label{fig:clinician_eval}
\end{figure}

We hypothesize that a generalist model can produce realistic, text-aligned images across medically and visually distinct modalities. To test this, we conducted a reader study where physicians assessed synthetic images generated by MediSyn across surgery and ophthalmology, see Figure \ref{fig:clinician_eval}.

To assess image realism, we tasked 5 surgeons and 5 ophthalmologists to identify any synthetic images in 102 laparoscopic cholecystectomy image or optical coherence tomography (OCT) image pairs, respectively \cite{foolrad}. There were an equal number of pairs containing 2 real images, 2 synthetic images, or 1 synthetic and 1 real image. On average, the surgeons achieved a recall of $50.59 \pm 25.81\%$ and a precision of $63.68 \pm 7.39\%$, while the ophthalmologists achieved a recall of $43.53 \pm 17.57\%$ and a precision of $62.39 \pm 12.79\%$. Together, these results demonstrate that physicians struggled to reliably identify synthetic images. Notably, for both medical specialties, the low recall indicates that many synthetic images went undetected. Moreover, the modest precision shows that a considerable proportion of images suspected to be synthetic were, in fact, real. This indicates that synthetic images are difficult to distinguish from real medical images and suggests that the synthetic images are highly realistic.

 Next, we conducted a blinded evaluation of text-alignment using a separate set of 132 synthetic and real images of laparoscopic cholecystectomy or OCT. The same surgeons and ophthalmologists were asked to determine whether each laparoscopic cholecystectomy or OCT image was of sufficient quality to classify surgical phase or retinal disease, and then classify the correct surgical phase or disease. In addition, we recorded their confidence level on a scale from 1 to 5 (5 being most confident) for each clinical classification. On average, the surgeons judged a marginally higher proportion of real images to be of sufficient quality for classification ($84.55 \pm 12.65\%$ vs $82.73 \pm 9.44\%$). For real images, they also achieved a marginally higher average classification accuracy ($90.30 \pm 3.95\%$ vs $89.39 \pm 1.86\%$) and confidence rating ($3.93 \pm 0.51$ vs $3.78 \pm 0.45$ out of 5). Similarly, on average, the ophthalmologists judged a slightly higher proportion of real images to be of sufficient quality for classification ($80.30 \pm 18.31\%$ vs $77.58 \pm 18.97\%$). However, for real images, they also achieved a lower average classification accuracy ($93.33 \pm 2.30\%$ vs $99.09 \pm 0.83\%$) and a slightly lower average confidence rating ($4.26 \pm 0.41$ vs $4.36 \pm 0.38$ out of 5). Together, these results demonstrate that a generalist model can produce synthetic images across distinct medical modalities that approach real images in both visual quality and conceptual complexity.

\begin{figure}[H] 
    \centering
    \includegraphics[width=\linewidth]
    {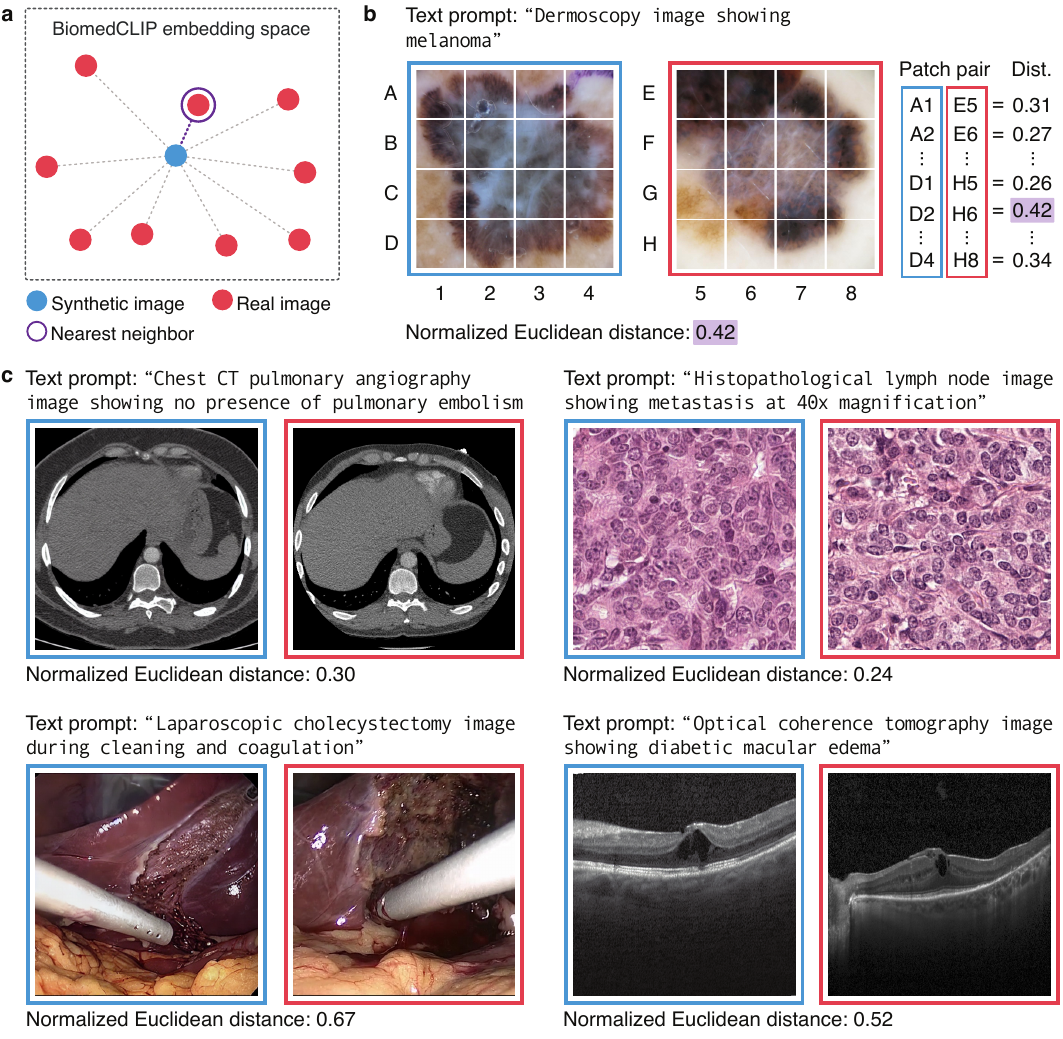} 
    \caption{\textbf{Assessment of training data reproduction.} \textbf{a.}  For each synthetic image, we found its nearest neighbor in the embedding space of BiomedCLIP's vision encoder. \textbf{b.} For each synthetic-real image pair, we calculated a normalized Euclidean distance between the two images. \textbf{c.} Additional examples of synthetic images alongside their nearest neighbor from the training dataset. }
    \label{fig:data_reproduction}
\end{figure}

\subsection{Synthetic images remain visually distinct from real patient data}
\label{privacy}
Recent work has shown that data memorization can occur in both text-guided latent diffusion models (LDMs) trained on natural images \cite{carlini2023extracting} and unconditional LDMs trained on medical images \cite{dar2025unconditional}. To assess whether MediSyn exhibits similar behavior, we first generated synthetic images across all 6 medical specialties included in our training dataset: (i) chest computed tomography (CT) images (radiology), (ii) dermoscopy images (dermatology), (iii) histopathological lymph node images (pathology), (iv) optical coherence tomography images (ophthalmology), (v) laparoscopic cholecystectomy images (surgery), and (vi) gastrointestinal endoscopy images (gastroenterology). We then identified their nearest neighbors in our training dataset based on cosine similarity in the embedding space of BiomedCLIP, a vision-language model for biomedicine \cite{biomedclip}. Finally, we calculated a normalized Euclidean distance for each synthetic-real image pair and considered distances less than or equal to 0.15 as instances of our model potentially reproducing our training data \cite{carlini2023extracting, Esser2024Scaling}, see Figure \ref{fig:data_reproduction}. We found that no radiology, pathology, ophthalmology, or gastroenterology image pair had a distance less than or equal to 0.15. However, some image pairs in dermatology and surgery contained distances less than or equal to 0.15. A secondary visual inspection confirmed that all synthetic images were distinct from their corresponding real images, see Supplementary Figure S3. 

To examine the sensitivity of our training data reproduction assessment to the distance threshold used, we repeated our evaluation for values ranging from 0.05 to 0.25 in increments of 0.05. Normalized Euclidean distances for all medical specialties were significantly greater than each threshold we examined, including 0.15 ($p < 0.001$ for each medical specialty). Moreover, visual inspection again confirmed that the synthetic images remained distinct from their corresponding real images across all thresholds, see Supplementary Figure S2.  See Table \ref{tab:EuclidDist} for a detailed breakdown of normalized Euclidean distances. In summary, these findings show that text-guided LDMs can consistently generate synthetic medical images visually distinct from real patient data, alleviating patient privacy concerns about image generative models in medicine. 

\begin{table}[t!]
    \centering
    \caption{Normalized Euclidean distance (NED) between synthetic images and their nearest neighbors in our training dataset. All distances are reported as mean $\pm$ 1 standard deviation. The columns labeled $\leq x$ denote the number (and percentage) of image pairs with NED at or below threshold $x$. }
    \small 
    \begin{tabular}{>{\raggedright\arraybackslash}m{0.16\linewidth} 
                    >{\raggedleft\arraybackslash}m{0.09\linewidth} 
                    >{\raggedleft\arraybackslash}m{0.06\linewidth} 
                    >{\raggedleft\arraybackslash}m{0.07\linewidth}
                    >{\raggedleft\arraybackslash}m{0.085\linewidth}
                    >{\raggedleft\arraybackslash}m{0.095\linewidth}
                    >{\raggedleft\arraybackslash}m{0.11\linewidth}
                    >{\raggedleft\arraybackslash}m{0.11\linewidth}}
        \toprule
        \textbf{Specialty} & \textbf{NED (-)} & \textbf{Image Pairs} & \textbf{$\leq$ 0.05} & \textbf{$\leq$ 0.10} &
        \textbf{$\leq$ 0.15} &
        \textbf{$\leq$ 0.20} &
        \textbf{$\leq$ 0.25} 
        \\
        \midrule
        Radiology        &$0.34 \pm 0.07$ & 1,000   & - & - & - & - & 28 (2.8\%)  \\
        Dermatology          &$0.27 \pm 0.10$ & 2,000 & - & 22 (1.1\%) & 173 (8.6\%) & 551 (27.6\%)  & 996 (49.8\%)  \\
        Pathology          & $0.31 \pm 0.05$ & 1,000   & - & - & - & 4 (0.4\%) & 174  (17.4\%)   \\
        Ophthalmology            & $0.39 \pm 0.12$ & 2,000  & - & - & - & 4 (0.2\%) & 94 (4.7\%) \\
        Surgery        & $0.38 \pm 0.11$ & 2,000   & - & - & 4 (0.2\%) & 46 (2.3\%) & 177 (8.8\%)     \\
        Gastroenterology         & $0.47 \pm 0.11$ & 2,000  & - & - & - & 6 (0.3\%) & 40 (2.0\%)    \\
        \bottomrule
    \end{tabular}
    
    \label{tab:EuclidDist}
\end{table}

\subsection{Generalist models can improve classifier performance in data-limited settings across multiple specialties }
\label{fine-tune}
\begin{figure}[t!] 
    \centering
    \includegraphics[width=\linewidth]
    {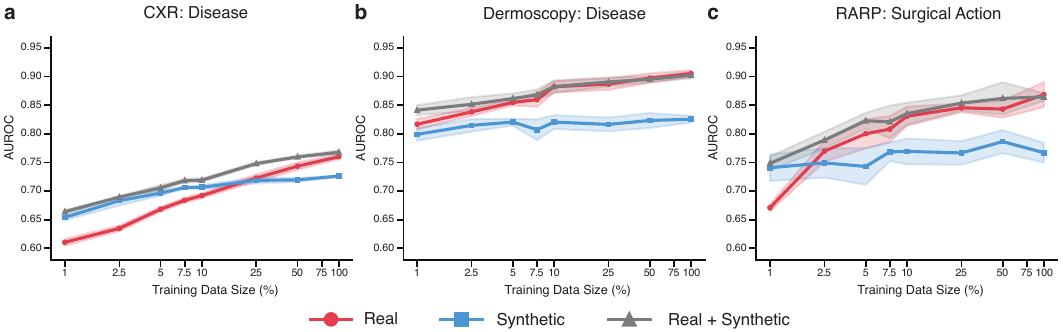} 
    \caption{\textbf{Performance of classifiers trained on either real data, synthetic data, or real data supplemented with synthetic data.} Macro-averaged test AUROC across five runs for ResNet-50 classifiers trained on varying proportions of the real data, the synthetic data generated by MediSyn, or the real data supplemented with the synthetic data. Classifiers were trained and evaluated on: \textbf{a}. chest X-ray images to classify diseases (multi-label classification), \textbf{b}. dermoscopy images to classify diseases (multi-class classification, and \textbf{c}. robot-assisted radical prostatectomy images to classify surgical actions (multi-class classification). All data presented as mean $\pm$ 1 standard deviation.}
    \label{fig:SyntheticClassifier}
\end{figure}

We posit that a generalist model can produce synthetic images suitable for training deep learning algorithms across multiple medical specialties, replacing or supplementing real data. To test this, we compared the performance of convolutional neural networks trained on real images, MediSyn-generated images, or real images supplemented with MediSyn-generated images, using 1\%, 2.5\%, 5\%, 7.5\%, 10\%, 25\%, 50\%, and 100\% of the full MediSyn training data, see Figure \ref{fig:SyntheticClassifier}. For radiology images, we found that in low-data settings ($\leq$ 25\% of the full training dataset), classifiers trained solely on synthetic images matched or outperformed those trained solely on real data. However, classifiers trained on real data supplemented with synthetic data outperformed those trained on solely real data, across all training ratios. In contrast, for dermatology and surgery, classifiers trained solely on synthetic data were outperformed by those trained solely on real data. However, in low-data settings ($\leq$ 50\% for surgery and $\leq$ 25\% for dermatology), training on real data supplemented with synthetic data produced the best results. These results highlight that a generalist model can produce synthetic images that improve the performance of DL algorithms in data-limited settings across multiple medical specialties . Detailed per-class performance is provided in Supplementary Tables S5–S7.

\section{Discussion}

Various groups have developed models to generate visually high-quality synthetic medical images, many of which are built upon the Stable Diffusion architecture \cite{Schutte2021, coyner2022synthetic, mcnulty2024synthetic, SHIFTS, ROENT, TEXTINV, dai2025improving, BUSGen}. However, these models are task-specific, restricted to a single medical specialty, imaging modality and anatomical region. In contrast, clinical practice encompasses a wide range of imaging types and applications, ranging from surgical action recognition to chest X-ray disease diagnosis \cite{sar-rarp, ROENT}. Consequently, synthetic data generation must support this diversity of medical imaging domains. Relying exclusively on a task-specific modeling approach would require the development, deployment, and maintenance of numerous task-specific image generative models, creating substantial barriers to scalability \cite{maintain_scale}. While Wang et al. \cite{MINIM} introduced a generative model trained across multiple imaging modalities, it is restricted to just 2 medical specialties and 4 imaging modalities, thereby limiting its scope. Moreover, similar to other medical image generative models \cite{MedSyn, dai2025improving, BUSGen}, its training images are not publicly available, thereby limiting transparency, reproducibility, and further development by the research community. In contrast, we train MediSyn exclusively on publicly available images covering 6 medical specialties and 10 imaging modalities. Notably, our generalist model produces images containing visually distinct objects such as soft tissues and surgical instruments in addition to radiological findings present in several existing models \cite{nwoye2024surgical, liu2025polypgen, ROENT}.  This allows for a single, unified model to potentially serve as a scalable source of synthetic data for training deep learning algorithms across a wide range of medical tasks. 

Generalist models find increasing use in biomedical imaging, with prior studies focusing on tasks such as multi-modal disease classification and radiological report generation \cite{biomedclip, medpalm-m}, reflecting a broader shift beyond task-specific approaches restricted to a singular medical imaging modality. MediSyn extends this generalist paradigm to text-guided image generation. Notably, prior work has demonstrated that conditional image generative models can function as zero-shot classifiers \cite{favero2025diffusionmedical, clark2023diffusion_zeroshot}, highlighting the potential of generalist, text-guided medical image generative models to support a wide range of downstream diagnostic tasks.

To evaluate the utility of our generalist approach, we performed five distinct evaluations of MediSyn. First, we found that generative training on visually diverse medical images does not degrade synthetic image quality. Specifically, MediSyn achieved competitive quantitative performance relative to task-specific models. For both dermoscopy and robot-assisted radical prostatectomy images, MediSyn achieved a lower FID but a higher MS-SSIM than the corresponding task-specific models. Conversely, for musculoskeletal X-ray images, MediSyn produced a higher FID but a lower MS-SSIM than the corresponding task-specific model. This possible tradeoff between image fidelity and diversity is further supported by the findings of Bluethgen et al. \cite{Bluethgen2024}, which demonstrate that models achieving lower FID scores did not necessarily exhibit lower MS-SSIM scores. 

Second, our generalist approach demonstrates substantially greater computational efficiency compared to a coordinated set of task-specific models for generating diverse medical images. MediSyn was more than twice as fast as the task-specific suite, with the primary overhead in the task-specific approach arising from repeatedly reloading model weights onto the GPU. These results are similar to prior work in ensemble modeling for medicine, where combining multiple deep learning models for clinical predictions incurs substantial computational costs \cite{ensemble}. 

Third, we found that a generalist model can produce realistic, text-aligned synthetic images across visually and medically distinct imaging modalities, as confirmed by expert physicians. In both specialties examined, physicians achieved low recall and modest precision, suggesting that synthetic images are difficult to distinguish from real ones. In addition, the surgeons achieved slightly higher classification accuracy on real images, whereas the ophthalmologists attained notably higher accuracy on synthetic images. This may suggest that the synthetic OCT images are visually less complex and varied, and that our model may generate less diverse outputs for certain imaging modalities. For comparison, Ktena et al. \cite{SHIFTS} reported that dermatologists achieved slightly lower accuracy when diagnosing synthetic skin lesion images compared to real ones. These differences may stem from variations in medical specialty and model architecture, with certain models producing more diverse outputs for specific imaging modalities than others.

Fourth, we found that our synthetic images were significantly visually different from their corresponding real patient data, alleviating widespread concerns surrounding data memorization in image generative models. While text-guided LDMs may memorize specific images, this is exacerbated by severely duplicated images in the training set \cite {carlini2023extracting} or images with inaccurate text prompts \cite{wang2024replication}. To address this, we hand-curated our training dataset to mitigate data duplication and ensured text prompts accurately reflected pre-existing class labels and image modality information. Additionally, although unconditional LDMs have been shown to memorize medical images \cite{dar2025unconditional}, our use of text conditioning introduces a semantic constraint that may encourage the model to generate images consistent with high-level descriptions rather than exact pixel-level replicas.

Finally, we found that a generalist model can produce images suitable for training deep learning algorithms across multiple medical specialties. Specifically, classifiers trained on real data supplemented with MediSyn-generated data outperformed those trained solely on real data in low-data settings. Conversely, those trained solely on MediSyn-generated data occasionally surpassed their real-only counterparts under similar conditions. This result is generally consistent with a similar, prior study on real and synthetic chest X-ray images \cite{Bluethgen2024}. Additionally, we observed that classifiers trained on real data supplemented with MediSyn-generated data surpass those trained solely on real data on certain clinical classes while underperforming on others. This suggests that diffusion models may be better at learning and synthesizing particular clinical conditions, which could later introduce unintended biases in models trained on their synthetic images \cite{Mittermaier2023}.

Our study is subject to several limitations. First, unlike natural image generative models \cite{zeroshot, Photorealistic}, our model lacks the zero-shot capability to generate quality images for types it has not explicitly encountered during training. For example, MediSyn is currently unable to generate chest MRI images, even though it has been trained on both MRI images and chest X-ray images. Second, similar to large language models \cite{asgari2025framework}, the model is still capable of hallucination, meaning it may generate images that are not clinically accurate, particularly when prompted with previously unseen disease descriptions. Third, while our model generates images distinct from real patient data, we currently cannot provide formal privacy guarantees. Ensuring such privacy protections without compromising visual quality would improve the reliability of synthetic data for clinical practice. Fourth, while our training dataset is extensive, it is inherently imbalanced, with certain modalities such as chest X-ray and surgical imaging substantially overrepresented relative to others. This imbalance may bias the model toward these dominant modalities and limit performance on underrepresented ones. Future work should focus on training strategies that address both data imbalance and inter-modality competition to ensure consistent performance across medical domains \cite{li2024sm3det}.

In conclusion, our findings reveal that generalist image generative models hold remarkable promise for accelerating algorithmic research and development in medicine. Specifically, these models provide a scalable approach to address data scarcity across numerous medical disciplines while maintaining rigorous standards of patient privacy. Overall, this work provides a robust foundation for the development of future generalist image generative models for medicine.

\section{Methods}\label{methods}
\subsection{Dataset curation and pre-processing}
We aggregated and pre-processed a collection of publicly available medical imaging data repositories spanning 6 medical specialties (Gastroenterology, Radiology, Pathology, Surgery, Dermatology, and Ophthalmology) and 10 imaging modalities (Computed Tomography, X-ray, Magnetic Resonance Imaging, Ultrasound, Endoscopy, Microscopy, Fundoscopy, Optical coherence tomography, Dermoscopy, and Clinical images). We divided each data repository into training, validation, and testing sets with an approximate 80:10:10 split at the patient level wherever possible. Each data repository was converted into an image-text format by converting modality information and existing class labels into descriptive text captions. Subsequently, we curated a medical dataset of 1,260,826 image-text pairs, with 1,014,781 pairs for training, 122,463 for validation, and 123,582 for testing. Please refer to Section 1.1 of the Supplementary Appendix for further details on data pre-processing. Further details on the data and their download links are provided in Supplementary Tables S1--S3. Finally, we detail how our curated dataset is utilized in Supplementary Figure S1. 

\subsection{Model architecture and training}
Similar to prior studies by Bluethgen et al. \cite{Bluethgen2024} and Wang et al. \cite{MINIM}, we evaluated the widely used Stable Diffusion version 1.4 (SD) \cite{Rombach_2022_CVPR} and its significantly larger variant, Stable Diffusion XL version 1.0 (SDXL) \cite{SDXL}, on our extensive dataset. SD and SDXL integrate three major components to generate a synthetic image conditioned on a text prompt: 
\begin{enumerate}
    \item \textbf{Variational Autoencoder}: The variational autoencoder (VAE) is an image generative model with an encoder-decoder architecture \cite{VAE}. First, the encoder condenses an image into a probabilistic distribution. Next, the decoder transforms a randomly selected sample from the distribution back to the original image. In both SD and SDXL, the encoder of the VAE is used exclusively in the training phase for encoding images into a lower-dimensional latent space where denoising operations occur. Conversely, the VAE decoder is reserved for image synthesis, where it transforms the denoised representations into the final output images.

     \item \textbf{Text Encoder(s)}: A text encoder is used to obtain a rich semantic representation of text which facilitates various language model tasks. SD uses a single text encoder extracted from CLIP ViT-L \cite{CLIP} for embedding text prompts. In contrast, SDXL employs two text encoders extracted from CLIP ViT-L and OpenCLIP ViT-bigG \cite{cherti2023reproducible}. Both individually process the conditioning text prompt, and their outputs are then concatenated along the channel axis.
    
    \item \textbf{Denoising U-Net}: The U-Net is a convolutional neural network originally designed for semantic segmentation \cite{UNet}. It features a contracting path of downsampling layers and a symmetric expansive path of upsampling layers. Due to its ability to produce outputs with the same dimensions as its inputs, DDPMs such as SD and SDXL have adopted the U-Net for noise prediction in images. Furthermore, text-conditioning in both SD and SDXL is enabled by the integration of a cross-attention mechanism between the intermediate layers of the U-Net. SD uses a U-Net with approximately 860 million parameters, while SDXL uses a significantly larger U-Net with approximately 2.6 billion parameters.
\end{enumerate}

All model components were initialized using weights from the HuggingFace Hub. For SD, the text encoder, variational autoencoder (VAE), and U-Net were initialized from the repository ``CompVis/stable-diffusion-v1-4." For SDXL, the text encoders and U-Net were initialized from the repository ``stabilityai/stable-diffusion-xl-base-1.0," while the VAE was initialized from the repository ``madebyollin/sdxl-vae-fp16-fix." We explored various distinct training procedures for each model, resulting in 6 total configurations. For both SD and SDXL, we first froze the VAE and fine-tuned both the CLIP ViT-L text encoder and U-Net. Second, we froze both the VAE and CLIP ViT-L text-encoder, and fine-tuned only the U-Net. Please note that for SDXL, the OpenCLIP ViT-bigG text encoder remained frozen for all training procedures due to computational constraints. For SD, we also explored replacing the CLIP ViT-L text encoder with BiomedCLIP's text encoder, a vision-language model for biomedicine \cite{biomedclip}. In this setup, we explored two approaches: jointly fine-tuning the U-Net and biomedical text encoder, and freezing the biomedical text encoder while fine-tuning only the U-Net. We fine-tuned all models for 15 epochs with an effective batch size of 128, using a constant learning rate of 5e-5 for SD and 1e-5 for SDXL. We used Bfloat16 floating point precision throughout training. To compute gradients, we calculated the Mean Squared Error (MSE) Loss between the predicted noise ($\hat{\epsilon}$) and its corresponding ground truth ($\epsilon$) \begin{figure}[h!]
\centering
\begin{minipage}{\textwidth}
\begin{equation}
\mathcal{L}_{\text{MSE}} = \frac{1}{H \times W} \sum_{i=1}^{H} \sum_{j=1}^{W} \left( \hat{\epsilon}_{i,j} - \epsilon_{i,j} \right)^2
\end{equation}
\end{minipage}
\end{figure}

Prompts were replaced with empty texts 10\% of the time to enable classifier-free guidance (CFG) \cite{CFG}, a method that allows a trade-off between image diversity and fidelity by controlling a hyperparameter scale. Higher values increase alignment to text prompts, while lower values lead to stronger mode coverage. Distributed training was conducted on either 4 NVIDIA H100 or 4 A100 graphics processing units (GPUs) (80GB VRAM) using the accelerate library (version 0.34.2) \cite{accelerate}, spanning roughly 2 days. We saved model checkpoints at the end of every epoch. SD with only the U-Net fine-tuned achieved the lowest Fréchet inception distance (FID) \cite{heusel2017gans} score, so this model configuration was designated as ``MediSyn" and used for subsequent evaluations \cite{ROENT}. Further details on the evaluation and analysis of the model configurations are provided in Sections 1.2 and 1.3 of the Supplementary Appendix.

\subsection{Model Evaluations}

\subsubsection{Assessment of synthetic image quality under diverse medical image training}
We aimed to assess whether generative training on visually diverse medical images degrades synthetic image quality. To this end, we measured and compared the image fidelity and diversity of MediSyn's outputs with those of task-specific models trained on datasets restricted to a single medical specialty, imaging modality, and anatomical region. Specifically, we fine-tuned the U-Net of Stable Diffusion version 1.4 (SD) \cite{Rombach_2022_CVPR} separately on the full set of 23,144 dermoscopy images, 32,151 musculoskeletal X-ray images, and 77,000 robot-assisted radical prostatectomy images contained in MediSyn's training dataset. To ensure a fair comparison, each task-specific model was fine-tuned for the same number of epochs as MediSyn. For each task-specific model, we randomly sampled 10,000 real images and their associated text prompts from its respective training dataset, and then generated 10,000 corresponding synthetic images using those prompts. Using each set of prompts, we also generated 10,000 corresponding synthetic images with MediSyn. To assess image fidelity, we computed the FID between each set of synthetic images and their corresponding real images \cite{Schutte2021, MINIM}. In addition, to assess image diversity, we randomly formed randomly 50,000 image pairs within each synthetic image set, and ensured that all 10,000 synthetic images were present in at least one pair. For each set of pairs, we calculated the multi-scale structural similarity index metric (MS-SSIM) \cite{MS-SSIM, ROENT} score for every pair and reported the mean and standard deviation.

\subsubsection{Assessment of computational efficiency against task-specific suite}

To assess the computational efficiency of our generalist approach, we measured and compared inference-time costs for diverse medical image generation using MediSyn versus a coordinated suite of task-specific models. We first sampled 10,000 text prompts from MediSyn’s training dataset in proportion to the underlying data distribution, and used these prompts to generate images with both MediSyn and the task-specific suite.

In the task-specific suite, each model was restricted to a single medical specialty, imaging modality, and anatomical region. We implemented a rule-based router that assigned each text prompt to its corresponding task-specific model. When successive prompts targeted different models, we emulated a model switch by loading the corresponding U-Net weights from disk into CPU memory and transferring them to the GPU to replace the currently active parameters, after which image generation proceeded as usual. Importantly, we did not train separate task-specific models for this experiment. Instead, we reused the same MediSyn U-Net weights and reloaded them at each sub-type transition to emulate switching between independent models. This design does not affect the validity of the comparison, as the computational cost of loading and transferring U-Net weights is independent of whether the weights originate from a generalist or task-specific model.

For both MediSyn and the task-specific suite, we repeated the experiment five times, each with a different random shuffling of the text prompt order. We first measured the time required to initialize the full SD pipeline (text encoder, variational autoencoder, and U-Net). For MediSyn, this initialization included an additional step of replacing the default SD U-Net weights with the MediSyn's fine-tuned weights. The task-specific suite did not require weight replacement at initialization, as the required task-specific model is determined by the first text prompt. We then measured image generation times for both approaches. For the task-specific suite, we additionally measured the time spent on prompt routing, loading task-specific U-Net weights into CPU memory, and transferring those weights to the GPU. All experiments were conducted using a single NVIDIA A100 GPU on the same compute node to ensure a fair comparison.

\subsubsection{Assessments of realism and text-alignment}
\label{subsubsec:visual-quality}

To evaluate whether a generalist model can produce realistic, text-aligned images across visually and medically distinct modalities, we conducted physician assessments in surgery and ophthalmology. For image realism, we tasked 5 surgeons and 5 ophthalmologists to conduct visual assessments of synthetic images from their specialties. Specifically, we asked them to identify synthetic data in 102 image pairs. These pairs contained 102 synthetic and 102 real images with an equal number of pairs containing 2 real images, 2 synthetic images, or 1 synthetic and 1 real image \cite{foolrad}. Each pair contained images from the same surgical phase or retinal disease class. Image pairing, order of images within each pair, and sequence of pairs shown were all randomized. Furthermore, all images were presented in the same sequence to each surgeon or ophthalmologist. We quantified physician performance using recall, defined as: 

\begin{equation}
    \text{Recall} = \frac{\text{No. Synthetic Images Correctly Identified} }{\text{Total No. of Synthetic Images}}
\end{equation}

as well as precision, defined as: 

\begin{equation}
    \text{Precision} = \frac{\text{No. of Synthetic Images Correctly Identified} }{\text{No. of Synthetic Images Identified}}
\end{equation}

Next, we tasked the same physicians to assess the text-alignment of our synthetic surgical and ophthalmological images. For a sequence of 132 images (66 real and 66 synthetic), we asked physicians to indicate whether image quality was sufficient for classification and to select the correct clinical label (surgical phase or retinal disease). In addition, we recorded their confidence level on a scale from 1 to 5 (5 being most confident) for each clinical classification. The images were presented in a randomized order, and physicians were blinded to the real or synthetic nature of each image. Furthermore, all images were presented in the same sequence to each surgeon or ophthalmologist.

The real images used in both assessments were randomly selected from the training dataset of MediSyn. For surgery, we used laparoscopic cholecystectomy images with phases ``\textit{Clipping and cutting}," ``\textit{Calot's triangle dissection}," and ``\textit{Cleaning and coagulation}." For ophthalmology,  we used optical coherence tomography (OCT) images with ``\textit{Normal}," ``\textit{Drusen}," and ``\textit{Diabetic macular edema}" labels.    

For each assessment, we generated 15,000 synthetic images for each of the 3 surgical phases or retinal diseases. For each real image used in each assessment, we identified its nearest neighbor within the same surgical phase or retinal disease from the corresponding set of 15,000 synthetic images. The nearest neighbor search was performed in the embedding space of BiomedCLIP's vision encoder based on cosine similarity. These nearest neighbors served as the synthetic images for the physician assessments, reducing the likelihood of mismatches in image complexity between the real and synthetic images.

\subsubsection{Assessment for training data reproduction}
\label{subsubsec:fine-tune}
To assess for training data reproduction, we performed pairwise comparisons between synthetic images generated by MediSyn and their nearest neighbors in our training dataset. First, we generated synthetic images across 6 medical specialties using text prompts derived from our original training data.

\begin{enumerate}
    \item \textbf{Radiology}: We used prompts corresponding to chest computed tomography (CT) images, constructed as either ``\textit{Chest CT pulmonary angiography image showing no presence of pulmonary embolism}," or ``\textit{Chest CT pulmonary angiography image showing pulmonary embolism.}" We generated 500 images from each of the two prompts.
    \item \textbf{Dermatology}: We used prompts describing dermoscopy images, constructed as ``\textit{Dermoscopy image showing \{disease\}}." We generated 500 images for each of the following diseases:  ``\textit{melanoma}," ``\textit{seborrheic keratosis},"  ``\textit{basal cell carcinoma}," and ``\textit{melanocytic nevus}." 
    \item \textbf{Pathology}: We used prompts describing histopathological lymph node images, constructed as either ``\textit{Histopathological lymph node image showing no presence of metastasis}," or ``\textit{Histopathological lymph node image showing metastasis}." We generated 500 images from each prompt.
    \item \textbf{Ophthalmology}: We used prompts describing optical coherence tomography images, constructed as ``\textit{Optical coherence tomography image showing \{disease\}}." We generated 500 images for each of the following classes: ``\textit{normal findings}," ``\textit{drusen}," ``\textit{choroidal neovascularization}," and ``\textit{diabetic macular edema}."
    \item \textbf{Surgery}: We used prompts describing laparoscopic cholecystectomy images, constructed as ``\textit{Laparoscopic cholecystectomy image during \{surgical phase\}}." We generated 500 images each for the following surgical phases: ``\textit{cleaning and coagulation}," ``\textit{Calot's triangle dissection}," ``\textit{gallbladder packaging}," and ``\textit{preparation}."
    \item \textbf{Gastroenterology}: We used prompts describing gastrointestinal endoscopy images, constructed as ``\textit{Gastrointestinal endoscopy image showing \{class state\}}." We generated 500 images each for the classes: ``\textit{polyp}," ``\textit{esophagitis}," ``\textit{ulcerative colitis}," and ``\textit{dyed and lifted polyp}." 

\end{enumerate}

Next, we used BiomedCLIP's vision encoder to extract feature representations for the synthetic images and real images in the training set. Following prior work\cite{carlini2023extracting}, which used CLIP to find nearest neighbors for synthetic natural images, we adopted a similar approach with BiomedCLIP to better align with the medical imaging domain. For each synthetic image, we then performed a nearest-neighbor search based on their cosine similarity \cite{Schutte2021}. We restricted the search to training images with the same medical specialty and image type. To quantitatively measure the similarity between pairs of synthetic and real images, we used a normalized, patch-based Euclidean distance in pixel space, 

\begin{figure}[H]
    \centering
    \begin{minipage}{\textwidth}
        \begin{equation}
        d = \frac{\left( \sqrt{ \sum_{i=1}^{N} \left( P_{1,i} - P_{2,i} \right)^2 } \right)}{\sqrt{(255^2) \cdot N}}
        \end{equation}
    \end{minipage}
\end{figure}

where $P_1$ refers to a patch of the real image, $P_2$ refers to a patch of the synthetic image, and $N$ represents the total number of pixels in each patch. This metric ranges from 0 to 1, with 0 indicating that the image patches are identical. For each real-synthetic image pair, we first divided each image into 16 non-overlapping $128\times128$ patches. Next, we calculated the maximum normalized Euclidean distance among all pairs of corresponding patches. They found that this approach mitigates any spurious similarities between images, such as those caused by a shared background color--a factor particularly relevant in medical imaging \cite{EchoMAE}. Pairs with distances $\leq 0.15$ are flagged as potential instances of training data reproduction, as established by Carlini et al \cite{carlini2023extracting, Esser2024Scaling}. To examine the sensitivity of this assessment to the chosen threshold, we repeat our analyses for threshold values between 0.05 and 0.25 in increments of 0.05. In addition, we visually examined pairs with distances below or at each threshold as a secondary check for data reproduction. 

\subsubsection{Assessment for algorithmic training using synthetic data}
We aimed to evaluate whether a generalist image generative model can replace or supplement real data across multiple medical disciplines. To this end, we conducted a series of experiments comparing the performance of classifiers within each discipline trained only on MediSyn-generated data, only on real data, or real data supplemented with an equal amount of MediSyn-generated data. We repeated this process with 1\%, 2.5\%, 5\%, 7.5\%, 10\%, 25\%, 50\%, and 100\% of the training set and approximately matched all types of datasets in class distribution. In addition, we validated and tested all classifiers against the same held-out validation and test sets from MediSyn, see Supplementary Figure S1. Our evaluation covered the following medical specialties represented in our training data: 

\begin{enumerate}
    \item \textbf{Radiology}: We focused on frontal, anterior-posterior (AP) chest X-ray images for a multi-label classification task targeting four diseases: cardiomegaly, edema, lung opacity, and pleural effusion. All chest X-ray images used in this experiment exclusively exhibited one or more of these diseases. The text prompts used to generate the synthetic images followed the structure: ``\textit{Frontal (AP) chest X-ray image showing \{diseases(s)\}}." We trained classifiers on 1\% (238 images), 2.5\% (583 images), 5\% (1,161 images), 7.5\% (1,737 images), 10\% (2,307 images), 25\% (5,705 images), 50\% (11,398 images), and 100\% (22,790 images) of the real training data, corresponding synthetic data, or the real data supplemented with the synthetic data, resulting in double the number of images at each respective percentage. Our validation and test sets of this classification task contained 2,897 and 2,872 images, respectively.
    \item \textbf{Dermatology}: We focused on dermoscopy images for a multi-class classification task targeting four diseases: basal cell carcinoma, melanocytic nevus, melanoma, and seborrheic keratosis. The text prompts used to generate the synthetic images followed the structure: ``\textit{Dermoscopy image showing \{disease\}}." We trained classifiers on 1\% (195 images), 2.5\% (492 images), 5\% (985 images), 7.5\% (1,478 images), 10\% (1,973 images), 25\% (4,937 images), 50\% (9,876 images), and 100\% (19,755 images) of the real training data, corresponding synthetic data, or the real data supplemented with the synthetic data, resulting in double the number of images at each respective percentage. Our validation and test sets of this classification task contained 2,516 and 2,460 images, respectively.
    \item \textbf{Surgery}: We focused on robot-assisted radical prostatectomy images for a multi-class classification task targeting four surgical actions: cutting the suture, picking up the needle, positioning the needle tip, and pushing the needle through the tissue. The text prompts used to generate the synthetic images followed the structure: ``\textit{Robot-assisted radical prostatectomy image of \{surgical action\}}." We trained classifiers on 1\% (530 images), 2.5\% (1,329 images), 5\% (2,659 images), 7.5\% (3,991 images), 10\% (5,324 images), 25\% (13,314 images), 50\% (26,629 images), and 100\% (53,261 images) of the real training data, corresponding synthetic data, or the real data supplemented with the synthetic data, resulting in double the number of images at each respective percentage. Our validation and test sets of this classification task contained 6,142 and 6,648 images, respectively.
\end{enumerate} 

We trained an ImageNet pre-trained ResNet-50 classifier \cite{he2015deep} on each combination of dataset type (e.g., real, synthetic) and data ratio (e.g., 1\%, 2.5\%). We used the models with the highest macro-averaged area under the Receiver Operating Characteristic (AUROC) values on the validation set and calculated their AUROC values on the corresponding test set.  This training and evaluation process was performed separately for each of the three medical specialties. All training and evaluation runs were repeated five times with different initialization seeds. We report the mean and standard deviation of the test AUROC across those runs.

\subsection{Statistical analyses}
\label{Stats}
 For each of the 6 medical specialties represented in the dataset of MediSyn, we conducted a one-sided, non-parametric Wilcoxon signed-rank test to assess whether the observed Euclidean distances from the real-synthetic image pairs were above each established threshold from 0.05 to 0.25. To control the family-wise error rate in performing multiple comparisons, we applied a Bonferroni correction. The performance of classifiers trained on synthetic data, real data, and real data supplemented with synthetic data was reported as the mean and standard deviation across five independent training and evaluation runs using different random seeds. Statistical analyses were performed using python (version 3.11.5).

 \section{Data Availability}
All datasets used in this study are publicly available and can be downloaded via the links provided in Supplementary Table S3. 

\section{Code Availability}
MediSyn was trained using the diffusers library (\url{https://github.com/huggingface/diffusers}).
The inference script and trained model weights are publicly available at \url{https://huggingface.co/hiesingerlab/MediSyn}. 

\section{Acknowledgments}
We would like to thank the Stanford Sherlock cluster for providing computational resources and support that contributed to these research results.

\section{Declarations}
This work was supported, in part, by awards from the National Heart, Lung, and Blood Institute (1R01HL157235-01A1) and the American Heart Association (25IPA1454136) to WH.

\subsection{Competing interests}
The authors declare no competing interests.

\subsection{Authors' contributions}
JC, MM, CZak, and WH designed the experiments. JC performed the data collection and pre-processing. JC performed the computational experiments. AD, AK, WH, MD, and AChoi performed the visual assessment of surgical images. EK, KW, CSZ, ER, and LA performed the visual assessment of optical coherence tomography images. JC, MM, and DK wrote the manuscript. ML, AChaud, and RS provided additional technical advice. WH supervised the work.

\clearpage
\bibliography{sn-bibliography}

\end{document}